\documentclass[letterpaper, 10 pt, conference]{ieeeconf}  

\IEEEoverridecommandlockouts                              

\usepackage{graphics} 
\usepackage{epsfig} 
\usepackage{mathptmx} 
\usepackage{times} 
\usepackage{amsmath} 
\usepackage{amssymb}  

\usepackage[utf8]{inputenc} 
\usepackage[T1]{fontenc}    
\usepackage[pagebackref=true,breaklinks=true,letterpaper=true,colorlinks,bookmarks=true]{hyperref}
\usepackage{url}            
\usepackage{booktabs}       
\usepackage{amsfonts}       
\usepackage{nicefrac}       
\usepackage{microtype}      
\usepackage{color}
\usepackage[table]{xcolor}
\usepackage{cleveref}
\usepackage{graphicx}
\usepackage{float}
\usepackage{multirow}
\usepackage{diagbox}

\usepackage{algorithm}
\usepackage{algpseudocode}
\usepackage{arydshln}
\usepackage{marvosym}
\usepackage{colortbl}
\usepackage{makecell}
\usepackage{multicol}
\usepackage{multirow}
\usepackage{listings}

\usepackage{adjustbox} 

\definecolor{dkgreen}{rgb}{0,0.6,0}
\definecolor{gray}{rgb}{0.5,0.5,0.5}
\definecolor{mauve}{rgb}{0.58,0,0.82}
\usepackage{tikz}

\title{\LARGE \bf
Distribution-Aware Continual Test-Time Adaptation for \\Semantic Segmentation 
}

\author{Jiayi Ni$^{1*}$, Senqiao Yang$^{1*}$, Ran Xu$^{1*}$, Jiaming Liu$^{1\dagger}$, Xiaoqi Li$^{1}$, Wenyu Jiao$^{2}$,  \\Zehui Chen$^{1}$, Yi Liu$^{3}$, Shanghang Zhang$^{1}$~\textsuperscript{\Letter}
\thanks{$^1$ Jiaming Liu, Jiayi Ni, Senqiao Yang, Xiaoqi Li, Ran Xu, Zehui Chen and Shanghang Zhang are with National Key Laboratory for Multimedia Information Processing, School of CS, Peking University.
$^2$ Wenyu Jiao is with University of Washington. $^3$ Yi Liu is with Baidu Inc. }
\thanks{ *: Equal Contribution: ni\_jiayi@stu.pku.edu.cn. $\dagger$: Project Leader: jiamingliu@stu.pku.edu.cn \textsuperscript{\Letter} 
Corresponding Author: shanghang@pku.edu.cn}
\thanks{Our project will be released at  \href{https://github.com/RochelleNi/DAT}{https://github.com/RochelleNi/DAT}}
}

\begin{document}
\maketitle

\begin{abstract}

Since autonomous driving systems usually face dynamic and ever-changing environments, continual test-time adaptation (CTTA) has been proposed as a strategy for transferring deployed models to continually changing target domains. However, the pursuit of long-term adaptation often introduces catastrophic forgetting and error accumulation problems, which impede the practical implementation of CTTA in the real world. Recently, existing CTTA methods mainly focus on utilizing a majority of parameters to fit target domain knowledge through self-training. Unfortunately, these approaches often amplify the challenge of error accumulation due to noisy pseudo-labels, and pose practical limitations stemming from the heavy computational costs associated with entire model updates. In this paper, we propose a distribution-aware tuning (DAT) method to make the semantic segmentation CTTA efficient and practical in real-world applications. DAT adaptively selects and updates two small groups of trainable parameters based on data distribution during the continual adaptation process, including domain-specific parameters (DSP) and task-relevant parameters (TRP). Specifically, DSP exhibits sensitivity to outputs with substantial distribution shifts, effectively mitigating the problem of error accumulation. In contrast, TRP are allocated to positions that are responsive to outputs with minor distribution shifts, which are fine-tuned to avoid the catastrophic forgetting problem. In addition, since CTTA is a temporal task, we introduce the Parameter Accumulation Update (PAU) strategy to collect the updated DSP and TRP in target domain sequences. We conducted extensive experiments on two widely-used semantic segmentation CTTA benchmarks, achieving competitive performance and efficiency compared to previous state-of-the-art methods.

\end{abstract}

\section{Introduction}



Continuous semantic segmentation represents a fundamental task in autonomous driving \cite{siam2018comparative, chi2023bev, yang2023lidar} and robotic manipulation \cite{li2023manipllm}. Assuming that both training and test data are under the same distribution, deployed models can achieve remarkable performance.
However, in the real world, this assumption is frequently violated due to the continually changing environments, resulting in a significant degradation in performance \cite{sun2022shift, sakaridis2021acdc}. 
Therefore, the concept of continual test-time adaptation (CTTA) has been introduced \cite{Wangetal2022}. Unlike traditional test-time adaptation, CTTA focuses on handling a sequence of distinct distribution shifts over time, rather than just a single shift.  In addition, due
to data privacy, the CTTA process renders the source data inaccessible, making it more challenging but
expanding its applicability \cite{liang2023comprehensive}.

\begin{figure}[t]
\includegraphics[width=0.47\textwidth]{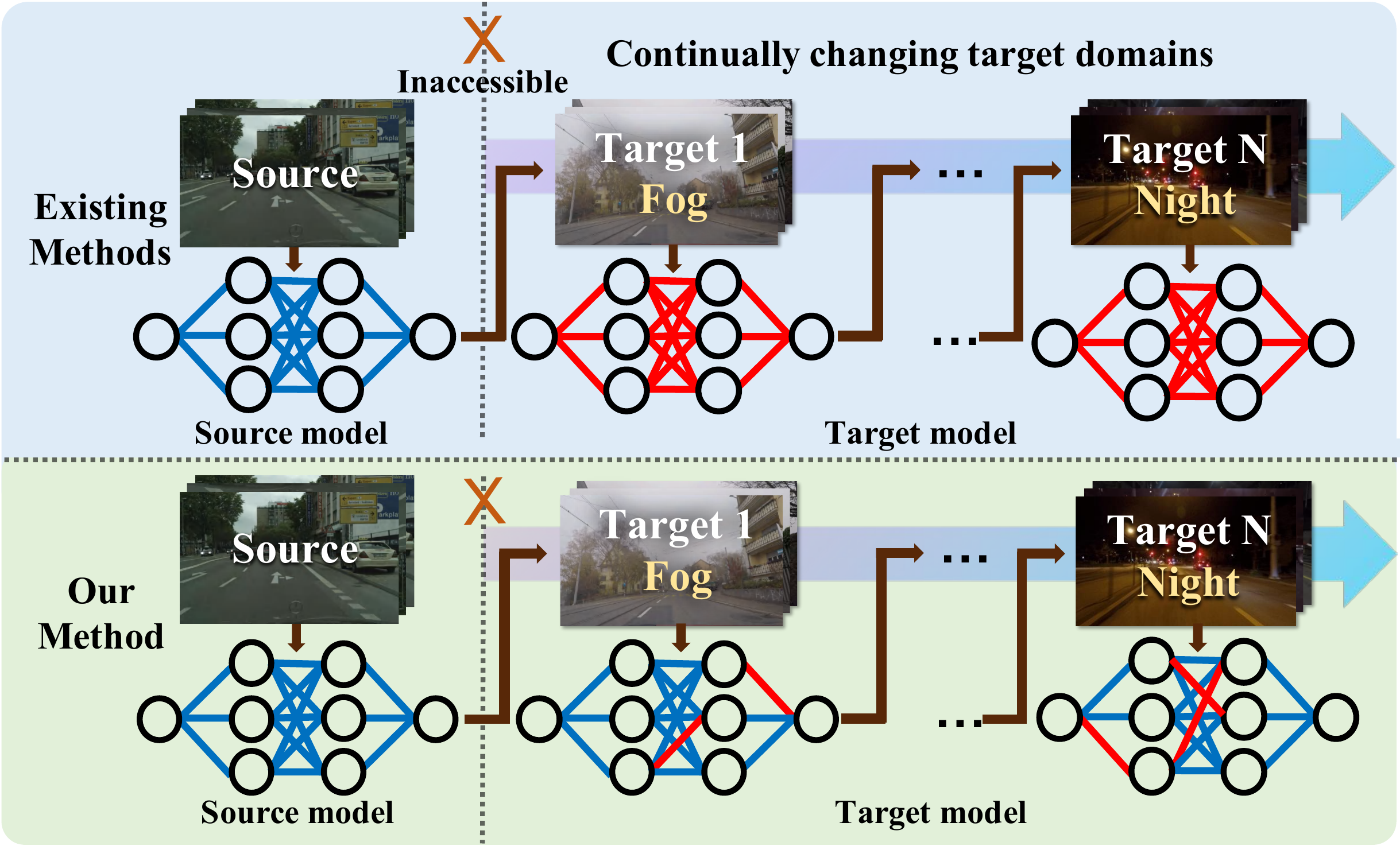}
\centering
\vspace{-0.3cm}
\caption{The CTTA problem aims to adapt the source model to continually changing target domains. Existing model-based methods focus on utilizing a majority of parameters to fit different target domain knowledge in continually changing environments. Differently, we propose a parameter-efficient model-based method named distribution-aware tuning (DAT), a novel paradigm for stable continual adaptation. DAT adaptively selects two small groups (e.g., 5\%) of trainable parameters based on the data distribution, enabling the simultaneous extraction of domain-specific and task-relevant knowledge. Instead of updating the entire model, DAT streamlines the continual adaptation process, enhancing its efficiency and potential for real-world applications. The red lines indicate the updated parameters.}
\label{fig:intro}
\vspace{-0.5cm}
\end{figure}
The pursuit of continual adaptation often introduces challenges such as catastrophic forgetting and error accumulation, hindering the practical implementation of CTTA in real-world applications. To address these issues, existing CTTA approaches have primarily employed model-based methods \cite{Wangetal2022, song2023ecotta} or parameter-efficient methods \cite{gan2022decorate,liu2023vida} to simultaneously extract target domain-specific and domain-invariant knowledge.
As shown in Fig. \ref{fig:intro}, the conventional model-based methods tend to update a majority of parameters for learning target domain knowledge through self-training. Unfortunately, these approaches frequently exacerbate the problem of error accumulation by introducing noisy pseudo-labels. Additionally, they pose practical limitations due to the substantial computational costs associated with updating the entire model.
Meanwhile, parameter-efficient methods leverage soft prompts \cite{jia2022visual} or adapters \cite{chen2022adaptformer} with limited trainable parameters to learn long-term domain-shared knowledge, leading to difficulties in preventing catastrophic forgetting.



In this paper, we propose a parameter-efficient model-based method named distribution-aware tuning (DAT), a novel paradigm for stable continual adaptation, shown in Fig. \ref{fig:intro}.
DAT does not require the addition of any prompts or adapters, avoiding disruption to the model's original structure. Instead, it leverages two small groups (around 5\%) of its own parameters to extract target domain-specific and task-relevant knowledge simultaneously.
Specifically, DAT first evaluates the degree of pixel-level distribution shift for each model output using the uncertainty mechanism \cite{gan2022cloud, roy2022uncertainty,
ovadia2019can}. At the same time, it divides the majority of pixels into two groups, representing significant distribution shifts and minor distribution shifts, respectively. DAT fine-tunes the selected domain-specific parameters (DSP) that exhibit sensitivity to outputs with substantial distribution shifts, effectively mitigating the problem of error accumulation. In contrast, task-relevant parameters (TRP) are allocated to positions that are responsive to outputs with minor distribution shifts, which are fine-tuned to avoid the catastrophic forgetting problem. 

Regarding the process of DSP and TRP selection, since CTTA is a task involving temporal sequences, we introduce the Parameter Accumulation Update (PAU) strategy to collect parameters. When processing a series of target domain samples, we only select a very small fraction of parameters (e.g., 0.1\%) for each sample to update and add these parameters to the parameter group. This process continues until the uncertainty value becomes relatively small, indicating that the collected parameter group can better understand current target domains.
After the accumulation stage, the accumulated parameter group (e.g., 5.0\%) will continue to be updated on the subsequent sequence data until encountering the next changing environment.



The main contributions can be outlined as follows:



\textbf{1)} We introduce an efficient distribution-aware tuning (DAT) method that adaptively selects two small groups of trainable parameters (e.g., 5.0\%), domain-specific parameters (DSP) and task-relevant parameters (TRP), based on the degree of distribution shifts. This approach addresses issues related to error accumulation and catastrophic forgetting.


\textbf{2)} For DSP and TRP selection, we employ the parameter accumulation update (PAU) strategy. When processing early target domain samples within the sequence, we select only a very small fraction of parameters (e.g., 0.1\%) for each sample and add these parameters to the updated parameter group until the distribution shift becomes relatively small.

\textbf{3)}
Our proposed method achieves competitive performance and efficiency compared to previous state-of-the-art methods on two CTTA benchmarks, Cityscape-ACDC \cite{sakaridis2021acdc} and SHIFT \cite{sun2022shift} continual datasets, showcasing its effectiveness in addressing the semantic segmentation CTTA problem.




\section{Related Work}

\textbf{Test-time adaptation (TTA)}, also known as source-free domain adaptation~\cite{liang2023comprehensive,gan2023cloud,wang2023cloud}, aims to adapt a source model to an unknown target domain distribution without the utilization of any source domain data.
These techniques mainly focus on unsupervised methods, such as self-training and entropy regularization, for the meticulous fine-tuning of the source model. Specifically, SHOT~\cite{Liangetal2020} distinguishes itself by concentrating on optimizing the feature extractor exclusively through information maximization and pseudo-labeling. AdaContrast~\cite{Chenetal2022}, while also leveraging pseudo labeling for TTA, incorporating self-supervised contrastive learning techniques.
Regarding model-based adaptation strategies, \cite{Boudiafetal2022} introduces a methodology involving the adjustment of the output distribution, showcasing its effectiveness in mitigating domain shift issues. Tent~\cite{DequanWangetal2021} updates the training parameters in the batch normalization layers through entropy minimization. This approach has prompted subsequent exploration in recent works~\cite{niu2023towards, yuan2023robust}, which continue to investigate the robustness of normalization layers.

\textbf{Continual Test-Time Adaptation (CTTA)} is a scenario where the target domain is dynamic, presenting increased challenges for traditional TTA methods.
Existing CTTA approaches have predominantly utilized either model-based methods \cite{Wangetal2022, song2023ecotta} or parameter-efficient methods \cite{gan2022decorate,liu2023vida} to concurrently address error accumulation and catastrophic forgetting issues.
For model-based methods, CoTTA \cite{Wangetal2022} enhances accuracy by employing weighted averaging and improved average prediction, all the while preventing catastrophic forgetting through random parameter recovery. Ecotta \cite{song2023ecotta}, on the other hand, introduces a meta-network to combat error accumulation by regularizing the outputs from both the meta-network and the frozen network.
Conversely, RMT \cite{dobler2023robust} addresses error accumulation through gradient analysis and introduces a symmetric cross-entropy loss tailored for CTTA. Additionally, \cite{li2023test} combines self-training from the supervised branch with pseudo labels from the self-supervised branch, providing a solution for the depth estimation CTTA problem.
For parameter-efficient methods, \cite{gan2022decorate} introduces the use of visual domain prompts specifically at the input level, extracting domain-specific and domain-agnostic knowledge. SVDP \cite{yang2023exploring} suggests the implementation of sparse visual prompts to address occlusion problems that arise from dense image-level prompts \cite{gan2022decorate}, thereby extracting sufficient local domain-specific knowledge.
\cite{liu2023vida} proposes ViDA to address error accumulation and catastrophic forgetting by leveraging low-rank and high-rank adapters to present different domain representations. 
Lastly, ADMA \cite{liu2023adaptive} introduces masked autoencoders to efficiently extract target domain knowledge and reconstructs hand-crafted invariant feature (e.g., Histograms of Oriented Gradients) to mitigate catastrophic forgetting.

\begin{figure*}[t]
\includegraphics[width=0.95\textwidth]{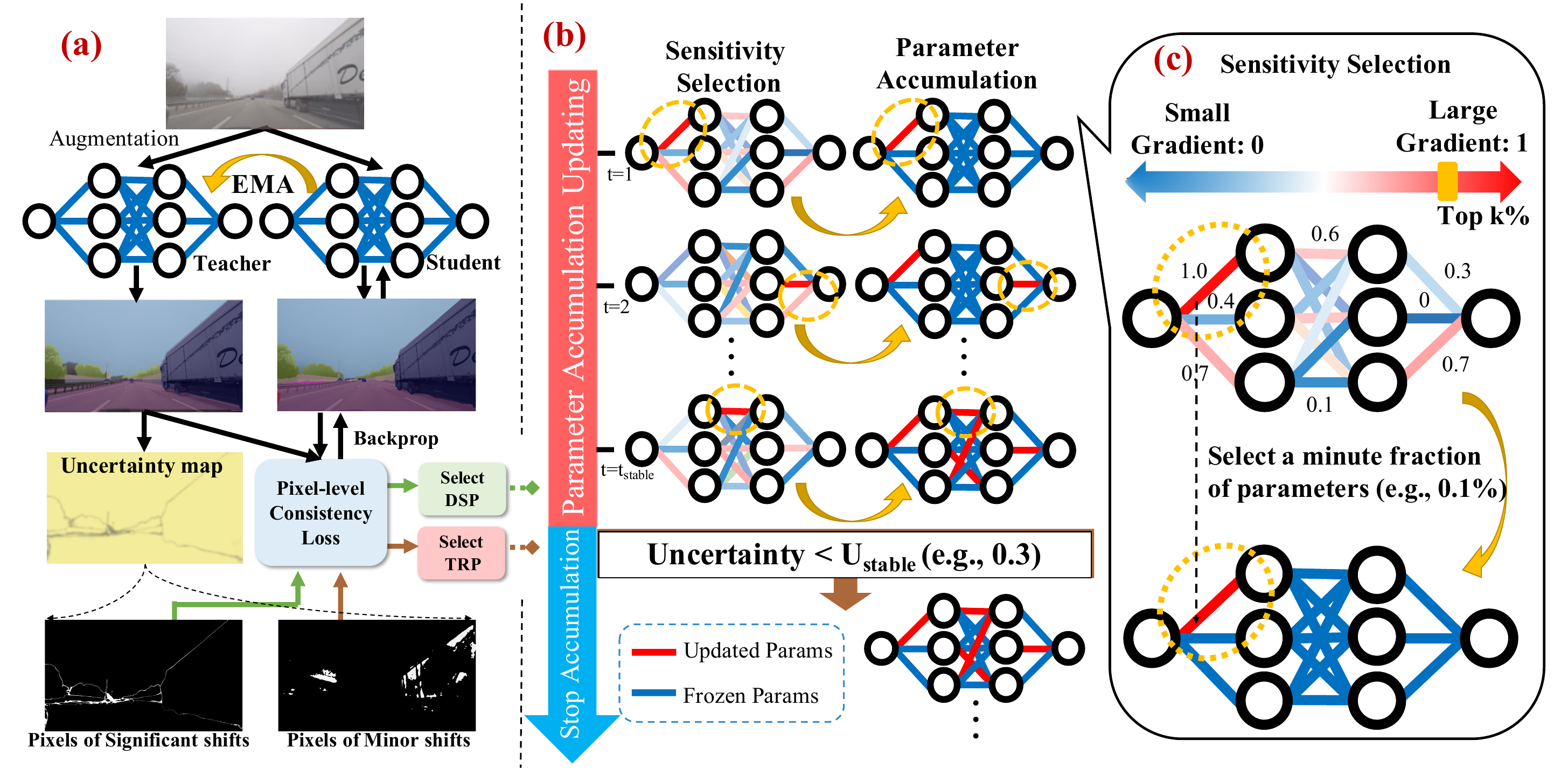}
\centering
\vspace{-0.35cm}
\caption{\textbf{The overall framework of distribution-aware tuning.} (a) In our approach, we implement a teacher-student framework to predict semantic segmentation maps, utilizing a pixel-wise consistency loss for optimization. Simultaneously, we obtain an uncertainty map from the teacher model to evaluate pixel-level distribution shifts. Based on the degree of these shifts, we selectively choose two small groups of trainable parameters (e.g., 5.0\%), namely domain-specific parameters (DSP) and task-relevant parameters (TRP).
(b) In the process of DSP and TRP selection, we employ the Parameter Accumulation Update (PAU) strategy to gather these parameters. When processing a series of target domain samples, we opt to select and update only a minute fraction of parameters (e.g., 0.1\%) for each sample, subsequently adding these parameters to the selected parameter group.
(c) We show the details of DSP and TRP selection, which involves identifying the most sensitive parameters for pixels with significant and minor distribution shifts, respectively.}
\label{fig:method}
\vspace{-0.5cm}
\end{figure*}

\section{Method}

\subsection{Preliminary}
\label{sec:3.1}
\textbf{Continual Test-Time Adaptation (CTTA)}
aims to adapt the source pre-trained model to constantly changing target domains during inference. A pre-trained model, $f_\theta$(x), with parameters $\theta$ trained on source data $\mathcal{X}_S$ and labels $\mathcal{Y}_S$, adapts online, refining predictions in a continually evolving target data without revisiting source data. 
The distributions of the target domains (i.e., $\mathcal{X}_{T_1}, \mathcal{X}_{T_2}, \dots, \mathcal{X}_{T_n}$) are in a state of constant change, and each sample within these target domains can only be utilized once.

\subsection{Overall framework and Motivation}
\label{sec:3.2}

\textbf{The overall framework}, as illustrated in Fig. \ref{fig:method} (a), begins with a teacher-student framework for predicting semantic segmentation maps, penalized by pixel-wise consistency loss. Simultaneously, we obtain an uncertainty map from the teacher model to assess pixel-level distribution shifts. Based on the uncertainty values, we divide all pixels from the student segmentation map into two groups: those with significant distribution shifts and those with minor shifts. We then use the same consistency loss to measure parameter gradient sensitivity for both groups, selecting the most sensitive parameters to represent domain-specific parameters (DSP) and task-relevant parameters (TRP), as depicted in Fig. \ref{fig:method} (c).
The Parameter Accumulation Update (PAU) strategy, shown in Fig. \ref{fig:method} (b), involves selecting a very small fraction of parameters (e.g., 0.1\%) for each sample and adding these parameters to the updated parameter group (represented by red lines). This process continues until the uncertainty value becomes relatively small, indicating that the parameter group is better suited to understand the current target domains. Note that, both DSP and TRP are selected through the PAU strategy during the continual adaptation process.

\textbf{Teacher-Student framework.}
Drawing from the insight that the mean teacher predictions often outperform standard models \cite{tarvainen2017mean}, we utilize a teacher model to enhance the accuracy of pseudo labels during continual domain adaptation. Furthermore, recognizing the robustness of the teacher-student framework in dynamic environments \cite{dobler2023robust}, we employ it to ensure stability within the target domains.

\textbf{Distribution-aware tuning.} 
Different from previous model-based CTTA methods, which update the majority of parameters for learning target domain knowledge, our approach aims to establish a parameter-efficient paradigm for stable continual adaptation. We achieve this by adaptively selecting two small groups of trainable parameters based on data distribution, allowing us to extract both target domain-specific and task-relevant knowledge. This approach allows us to update only the most crucial parameters for target domain samples, rather than updating the entire model, significantly enhancing the efficiency and practicality of CTTA in real-world applications.
Different from previous parameter-efficient CTTA methods, our approach does not require the additional module of prompts or adapters, preserving the model's original structure. Instead, we simply fine-tune two small groups of its existing parameters (DSP and TRP) to simultaneously avoid catastrophic forgetting and error accumulation problems.


\subsection{Distribution Aware Tuning}
\label{sec:3.3}

\textbf{Uncertainty Scheme} is an effective method for quantifying the unreliability of the model's predictions under distribution shift~\cite{liu2023vida, roy2022uncertainty, ovadia2019can}. Therefore, we utilize pixel-level uncertainty values to indirectly measure the degree of distribution shift.
To derive uncertainty values, we employ MC Dropout ~\cite{gal2016dropout} on the prediction head of the teacher model, facilitating multiple forward propagations to obtain distinct sets of segmentation maps ($m$ sets) for each sample. 
Moreover, we can directly employ multiple sets of augmented inputs with the teacher model to produce $m$-set segmentation maps for each sample.
We compute the uncertainty value for a given input. This value is defined as follows:

\begin{equation}
\mathcal{U} (\widetilde{x}_j) =  \left( \frac{1}{m} \sum_{i=1}^m \|p_i(\widetilde {y_j}|\widetilde {x_j}) - \mu \|^2 \right) ^{\frac{1}{2}}
\label{eq:mc}
\end{equation}

Where $p_i(\widetilde{y_j}|\widetilde{x_j})$ denotes the predicted probability of output $\widetilde{y_j}$ based on the input pixel $\widetilde{x_j}$ during the $i^{th}$ iteration of forward propagation; $\mu$ signifies the average prediction over $m$ iterations for $\widetilde{x_j}$. Therefore, $\mathcal{U} (\widetilde{x_j})$ represents the uncertainty regarding pixel-wise model input.

Hence, we can utilize the uncertainty value to classify the pixels of the output. The formulation is as follows:

\begin{equation}
\scriptstyle
   \left\{
    \begin{array}{l}
            \widetilde{x_j} \in \mathcal{G}_h
, \quad \mathcal{U} (\widetilde{x_j}) \ge \Theta_h\\  
             \widetilde{x_j} \in \mathcal{G}_l, \quad \mathcal{U} (\widetilde{x_j}) < \Theta_l
        \end{array}
        \right.
\end{equation}
The threshold value of uncertainty is denoted as $\Theta$, with $\Theta_h$ and $\Theta_l$ empirically set to 0.99 and 0.78. $\mathcal{G}_h$ and $\mathcal{G}_l$ represent the pixel groups with high and low uncertainty value.

\textbf{Domain-Specific Parameters (DSP).}
After calculating the uncertainty of each pixel, we select those with significant uncertainty values, indicating substantial distribution shifts.
Subsequently, we employ these pixels to select domain-specific parameters. This is specifically achieved by utilizing a binary mask to compute the loss and perform backpropagation exclusively for these pixels. Following this, we choose parameters with higher gradient values as Domain-Specific parameters. These domain-specific parameters exhibit sensitivity
to outputs with substantial distribution shifts. By updating these domain-specific parameters, they effectively capture domain-specific knowledge in the target domain, thus efficiently mitigating the issue of error accumulation.

\textbf{Task-Relevant Parameters (TRP).}
Apart from fine-tuning domain-specific parameters (DSP) to prevent error accumulation, task-relevant parameters (TRP) are selected to mitigate catastrophic forgetting. 
Using the identical procedure as described previously, we select pixels with lower uncertainty values, signifying minimal domain shifts and yielding more dependable segmentation outcomes.
Similarly, we only back-propagate through these pixels with lower uncertainty and select the parameters with the highest gradient as Task-Relevant Parameters, which are more focused on the task itself and are not affected by domain shifts.

Overall, based on the distribution of each sample, we select domain-specific parameters and task-relevant parameters. Then, we add them in the selected parameter group and utilize the PAU method to continually update these parameters.

\subsection{Parameter Accumulation Update}
\label{sec:3.4}
For the basic parameter update strategy, we employ Eq. \ref{eq:loss} to update the DSP and TRP of student model ($\mathcal{S}_{target}$) and leverage the widely-used exponential moving average (EMA) to update the corresponding parameters of teacher model ($\mathcal{T}_{mean}$), ensuring maximal model plasticity. The EMA process can be formulated as:
\begin{equation}
 \mathcal{T}_{mean}^{t} = \alpha \mathcal{T}_{mean}^{t-1} + (1-\alpha) \mathcal{S}_{target} ^{t}
\label{eq:ema}
\end{equation}
In this equation, t represents the time step (EMA update every frame), and $\alpha$ is set to 0.999, as referenced in \cite{AnttiTarvainenetal2017}.
When $t = 0$, representing the initial time step, we use the pre-trained model from the source domain to initialize the weights of both the teacher and student models.

Regarding the process of DSP and TRP selection, since CTTA and autonomous driving systems are both tasks involving temporal sequences, we introduce the Parameter Accumulation Update (PAU) strategy in addition to the existing EMA parameter updating approach. Specifically, when processing a series of target domain samples, for each individual image, we select Domain-Specific parameters and Task-Relevant Parameters based on their uncertainty. It's important to note that, to ensure the efficiency of the model, we only choose a very small fraction of parameters (e.g., 0.1\%) for each sample to update, and we add these selected parameters to the parameter group. 
This iterative process persists until the $100^{th}$ frame of the target domain sample, at which point we empirically determine that the accumulated parameter group is capable of fitting the current target domain. 
If the target domain dataset has less than 100 samples, the iterative process will encompass the entire dataset.
After the accumulation stage, the accumulated parameter group will continue to be updated on the subsequent sequence data. When encountering the next target domain, we reset all accumulated positions of DSP and TRP, starting a new round of parameter accumulation.


\subsection{Loss Functions}
\label{sec:3.5}
Following prior CTTA research~\cite{Wangetal2022}, the optimization objective is a consistency loss to minimize the discrepancy between outputs of teacher and student model, which is a cross-entropy loss calculated at the pixel level:
\begin{equation}
 \mathcal{L}_{con}(\widetilde {x}) = -
 \frac{1}{H \times W} \sum_{w,h}^{W,H}
 \sum_c^C \widetilde{y}_t(w,h,c) \log \hat{y}_t(w,h,c)
\label{eq:loss}
\end{equation}
In this equation, $\widetilde {x}$ is input, $\widetilde{y}_t$ is pseudo labels generated by the teacher model and $\hat{y}_t$ is student model output, whereas $C$ denotes the number of categories, and $H$ and $W$ refer to the height and width of images, respectively. 

\begin{table*}[htb]
\caption{\label{dg benchmark} \textbf{Performance comparison for Cityscape-to-ACDC CTTA.} During testing, we sequentially evaluate the four target domains three times. Mean denotes the average mIoU score. Gain quantifies the method's improvement compared to the Source model.}

\vspace{-0.3cm}
\centering
\setlength\tabcolsep{2pt}
\begin{adjustbox}{width=1\linewidth,center=\linewidth}
\begin{tabular}{c|c|ccccc|ccccc|ccccc|c|c }
\hline

\multicolumn{2}{c|}{Time}     & \multicolumn{15}{c}{$t$ \makebox[10cm]{\rightarrowfill} }                                                                              \\ \hline
\multicolumn{2}{c|}{Round}          & \multicolumn{5}{c|}{1}    & \multicolumn{5}{c|}{2}     & \multicolumn{5}{c|}{3}  & \multirow{2}{*}{Mean$\uparrow$}   & \multirow{2}{*}{Gain}  \\ \cline{1-17}
Method & REF & Fog & Night & Rain & Snow & Mean$\uparrow$ & Fog & Night & Rain & Snow  & Mean$\uparrow$ & Fog & Night & Rain & Snow & Mean$\uparrow$ & \\ \hline
Source & NIPS2021 \cite{xie2021segformer}  &69.1&40.3&59.7&57.8&56.7&69.1&40.3&59.7& 	57.8&56.7&69.1&40.3&59.7& 57.8&56.7&56.7&/\\ 
TENT & ICLR2021 \cite{wang2020tent}  &69.0&40.2&60.1&57.3&56.7&68.3&39.0&60.1& 	56.3&55.9&67.5&37.8&59.6&55.0&55.0&55.7&-1.0\\ 
CoTTA & CVPR2022\cite{Wangetal2022}  &70.9&41.2&62.4&59.7&58.6&70.9&41.1&62.6& 	59.7&58.6&70.9&41.0&62.7&59.7&58.6&58.6&+1.9\\ 
DePT & ICLR2023\cite{gao2022visual} 
&71.0&40.8&58.2&56.8&56.5&68.2&40.0&55.4&53.7& 54.3&66.4&38.0&47.3&47.2&49.7&53.4&-3.3\\
VDP & AAAI2023\cite{gan2022decorate}  &70.5&41.1&62.1&59.5&  58.3    &70.4&41.1&62.2&59.4& 58.2     & 70.4&41.0&62.2&59.4& 58.2   &  58.2 & +1.5\\
\cellcolor{lightgray}\textbf{DAT} &\cellcolor{lightgray}\textbf{ours} &\cellcolor{lightgray}\textbf{71.7}&\cellcolor{lightgray}\textbf{44.4}&\cellcolor{lightgray}\textbf{65.4}&\cellcolor{lightgray}\textbf{62.9}&\cellcolor{lightgray}\textbf{61.1}& 
 \cellcolor{lightgray}\textbf{71.6}&\cellcolor{lightgray}\textbf{45.2}&\cellcolor{lightgray}\textbf{63.7}&\cellcolor{lightgray}\textbf{63.3}&\cellcolor{lightgray}\textbf{61.0} 
 &\cellcolor{lightgray}\textbf{70.6}&\cellcolor{lightgray}\textbf{44.2}&\cellcolor{lightgray}\textbf{63.0}&\cellcolor{lightgray}\textbf{62.8}&\cellcolor{lightgray}\textbf{60.2}      &\cellcolor{lightgray}\textbf{60.8} 
 &\cellcolor{lightgray}+\textbf{4.1}\\\bottomrule
 \hline
\end{tabular}
\end{adjustbox}
\vspace{-0.3cm}
\label{tab:CTTA}
\end{table*}
\section{Experiments}
In Sec~\ref{sec:4.1}, we provide the details of the task settings for continual test-time adaptation (CTTA), as well as a description of the datasets. In Sec~\ref{sec:4.2}, we compare our method with other baselines~\cite{xie2021segformer, wang2020tent, Wangetal2022, gao2022visual} on two commonly used CTTA benchmarks. 
Comprehensive ablation studies are conducted in Sec~\ref{sec:4.3}, which investigate the impact of each component.

\subsection{Datasets and Implementation Details}
\label{sec:4.1}

\textbf{Cityscapes-to-ACDC} refers to a semantic segmentation task specifically designed for cross-domain learning in autonomous driving. The source model used is an off-the-shelf pre-trained segmentation model originally trained on the Cityscapes dataset~\cite{cordts2016cityscapes}. 
The ACDC dataset~\cite{sakaridis2021acdc} contains images collected in four different, previously unseen visual conditions: Fog, Night, Rain, and Snow. In the CTTA scenario, we replicate the sequence of target domains (Fog→Night→Rain→Snow) multiple times to simulate environmental changes in real-life driving scenarios.

\textbf{SHIFT} is an expansive synthetic driving video dataset that simulates real-world autonomous driving scenarios ~\cite{sun2022shift}. This dataset consists of two parts: a discrete set containing 4,250 sequences and a continuous set with an additional 600 sequences. 
In this configuration, the source model pre-trains on the discrete datasets and tests on the continuous validation dataset. The continuous set is designed to simulate gradual transitions between different environments, simulating real-world scenarios such as the continual transition from clear daytime conditions to nighttime in a validation sequence.

\textbf{Implementation Details.} We follow the basic implementation details \cite{Wangetal2022, gan2022decorate} to set up our semantic segmentation experiments. Specifically, in the Cityscapes-to-ACDC scenario, we employ the Segformer-B5 ~\cite{xie2021segformer} as our segmentation model and down-sample the original image size of 1920x1080 to 960x540, which serves as network input.
For the SHIFT dataset, we follow the setup details in~\cite{sun2022shift} and also employ the Segformer-B5. And we downsample the original image size to 800x500 during CTTA.
For both of these two datasets, we evaluate our predictions under the original resolution. 
We use the Adam optimizer~\cite{kingma2014adam} with $(\beta_1, \beta_2) = (0.9, 0.999)$, a learning rate of 3e-4 and batch size 1 for all experiments.
Similar to \cite{Wangetal2022}, we apply a range of image resolution scale factors [0.5, 0.75, 1.0, 1.25, 1.5, 1.75, 2.0] for the augmentation method in the teacher model.
No more than 10\% parameters of the model are updated.
All experiments are conducted on NVIDIA V100 GPUs.

\subsection{Effectiveness}
\label{sec:4.2}
\begin{figure*}[t]
\includegraphics[width=0.95\textwidth]{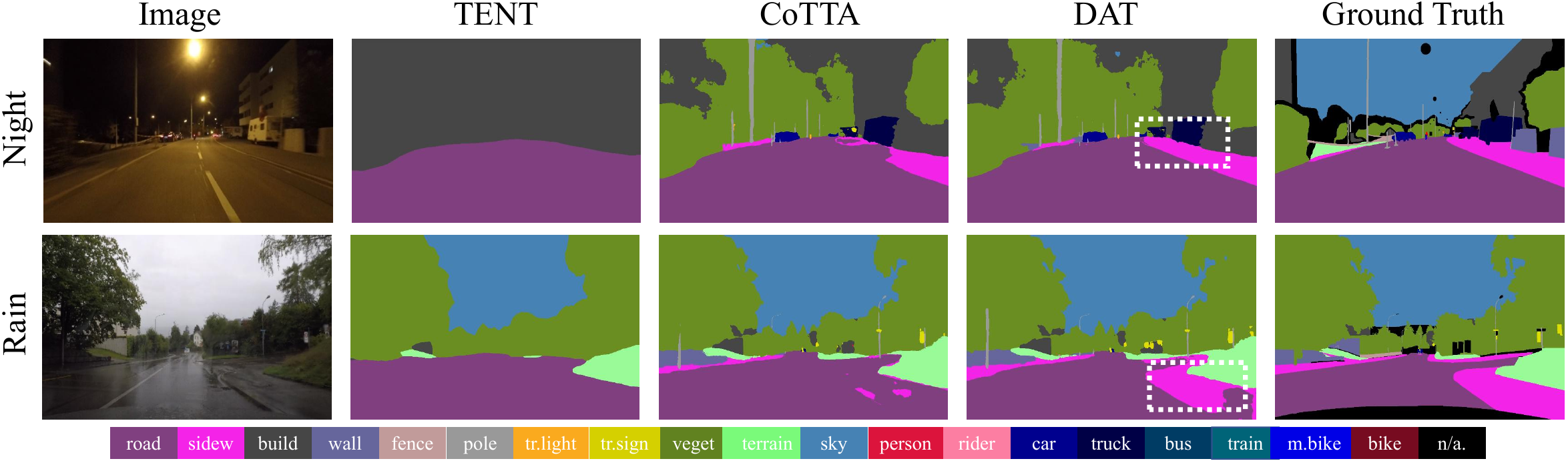}
\centering
\vspace{-0.2cm}
\caption{ Qualitative comparisons on the ACDC dataset. Our method could better segment different pixel-wise classes such as shown in the white box.}
\label{fig: vis}
\vspace{-0.3cm}
\end{figure*}

\textbf{Cityscapes-to-ACDC.} 
Given the source model pre-trained on Cityscapes, we conduct CTTA on ACDC dataset, which contains four severe weather that occur sequentially during the test time. As shown in Table~\ref{tab:CTTA}, the source model's mean Intersection over Union~(mIoU) is only 56.7\% when directly testing the source model on target domains in three rounds.
In contrast, our method consistently outperforms all previous methods, delivering a 4.1\% and 2.6\% improvement over the source model and CoTTA, respectively. 
Unlike TENT~\cite{wang2020tent}, which updates only the normalization layer, our DAT selects task-relevant parameters for fine-tuning, strategically allocating them to positions responsive to outputs with minor distribution shifts. Notably, TENT exhibits significant catastrophic forgetting, with mIoU dropping from 56.7 to 55.0. In contrast, our DAT does not exhibit this issue, underscoring the effectiveness of fine-tuning task-relevant parameters~(TRP) in mitigating catastrophic forgetting problems.
In comparison to CoTTA~\cite{Wangetal2022}, our method has significant improvement across most domains. This showcases our method's capacity to extract domain knowledge by fine-tuning domain-specific parameters, enabling rapid adaptation to target domains, without fine-tuning redundant parameters.
In conclusion, our DAT can effectively extract diverse domain knowledge while concurrently avoiding catastrophic forgetting simultaneously.
Furthermore, we present the qualitative results in Fig.~\ref{fig: vis}. Our method correctly distinguishes the sidewalk from the road.
\begin{table}[t]
\caption{\label{tab:Val} \textbf{Performance comparison for SHIFT dataset's continuous validation set.}  ACC denotes the average accuracy score.}
\centering
\vspace{-0.3cm}
\setlength\tabcolsep{1pt}
\begin{adjustbox}{width=0.99\linewidth,center=\linewidth}
\begin{tabular}{c|cc|cc|cc|c }
\hline
\multicolumn{1}{c|}{Scenarios}   & \multicolumn{2}{c|}{$Daytime - Night$}    & \multicolumn{2}{c|}{$Clear - Foggy$}     & \multicolumn{2}{c|}{$Clear - Rainy$}  &Mean    \\ \cline{1-7}
Method & mIoU & ACC & mIoU & ACC  & mIoU & ACC &mIoU  \\ \hline
Source  & 64.22 & 71.84 & 61.51  &  69.87 & 66.14 & 73.87 & 64.01 \\ 
TENT  & 64.21 & 71.67  &61.65 & 69.76  &  66.11 & 73.68 &  63.99\\
CoTTA  & 68.06 & 74.18 & 64.78 & 71.41 & 69.92 & 76.29 & 67.61  \\
\cellcolor{lightgray}\textbf{Ours} & \cellcolor{lightgray}\textbf{68.63} & \cellcolor{lightgray}\textbf{75.34} &\cellcolor{lightgray}\textbf{65.31} & \cellcolor{lightgray}\textbf{72.68} & \cellcolor{lightgray}\textbf{70.56} & \cellcolor{lightgray}\textbf{77.34}& \cellcolor{lightgray}\textbf{68.17}\\
 \hline\end{tabular}
\end{adjustbox}
\vspace{-0.5cm}
\label{tab:shift}
\end{table}

\textbf{SHIFT.} In order to verify the generalization of our proposed method, we conduct experiments on the autonomous driving CTTA datasets, specifically the SHIFT dataset~\cite{sun2022shift}. Using a source model pre-trained on SHIFT's discrete datasets, we conduct CTTA experiment on the gradually evolving SHIFT continuous validation set. As presented in Table~\ref{tab:shift}, the mIoU consistently demonstrates our method's superior performance compared to previous techniques across all three scenarios. Notably, in the most challenging scenario (from clear to foggy), our approach exhibited a significant enhancement over CoTTA, with mIoU increasing from 64.78 to 65.31 and mACC improving from 71.41 to 72.68. 
It's worth noting that although our approach has shown improvement compared to the previous method, the short length of each sequence in the SHIFT dataset and the strong dependency between the data's each frame make it challenging to fully demonstrate our method's advantage in preventing catastrophic forgetting.

\subsection{Ablation Study}
\label{sec:4.3}
In this subsection, we evaluate the contribution of each component in the Cityscape-to-ACDC scenario. 

\textbf{Effectiveness of each component} 
As presented in Table~\ref{tab:ablation} $Ex_{2}$, by introducing fine-tuning the domain-specific parameter (DSP), we observe that the mIoU increases by 3.5\%. The outcome underscores the efficacy that fine-tuning domain-specific parameters can effectively mitigate the problem of error accumulation.
As shown in $Ex_{3}$, TRP achieves further 3.6\% mIoU improvement since the task-relevant parameters (TRP) are allocated to
positions that are responsive to outputs with minor distribution
shifts, which are fine-tuned to avoid the catastrophic forgetting problem.
$Ex_{4}$ demonstrates the comprehensive integration of all components, resulting in a total mIoU improvement of 4.4\%. This underscores the effectiveness of our proposed method and illustrates that all components can collaboratively address the semantic segmentation CTTA problem in autonomous driving.
\begin{table}[!tb]
\caption{\textbf{Ablation: Contribution of each component. 
}}
\vspace{-0.3cm}
\centering
\setlength\tabcolsep{10.0pt}
\renewcommand\arraystretch{1.2}
\begin{tabular}{c|cc|c | c}
\hline
 & DSP & TRP   & mIoU$\uparrow$  & Gain 
\\\hline
$Ex_{1}$  &   & &56.7 & / \\ 
$Ex_{2}$     &\checkmark &   & 60.2 & +3.5\\
$Ex_{3}$     & &\checkmark     &60.3 & +3.6\\
$Ex_{4}$ &\checkmark  &\checkmark & 61.1 & +4.4\\
\hline
\end{tabular}
\vspace{-0.3cm}
\label{tab:ablation}
\end{table}

\begin{figure}[t]
\includegraphics[width=0.47\textwidth]{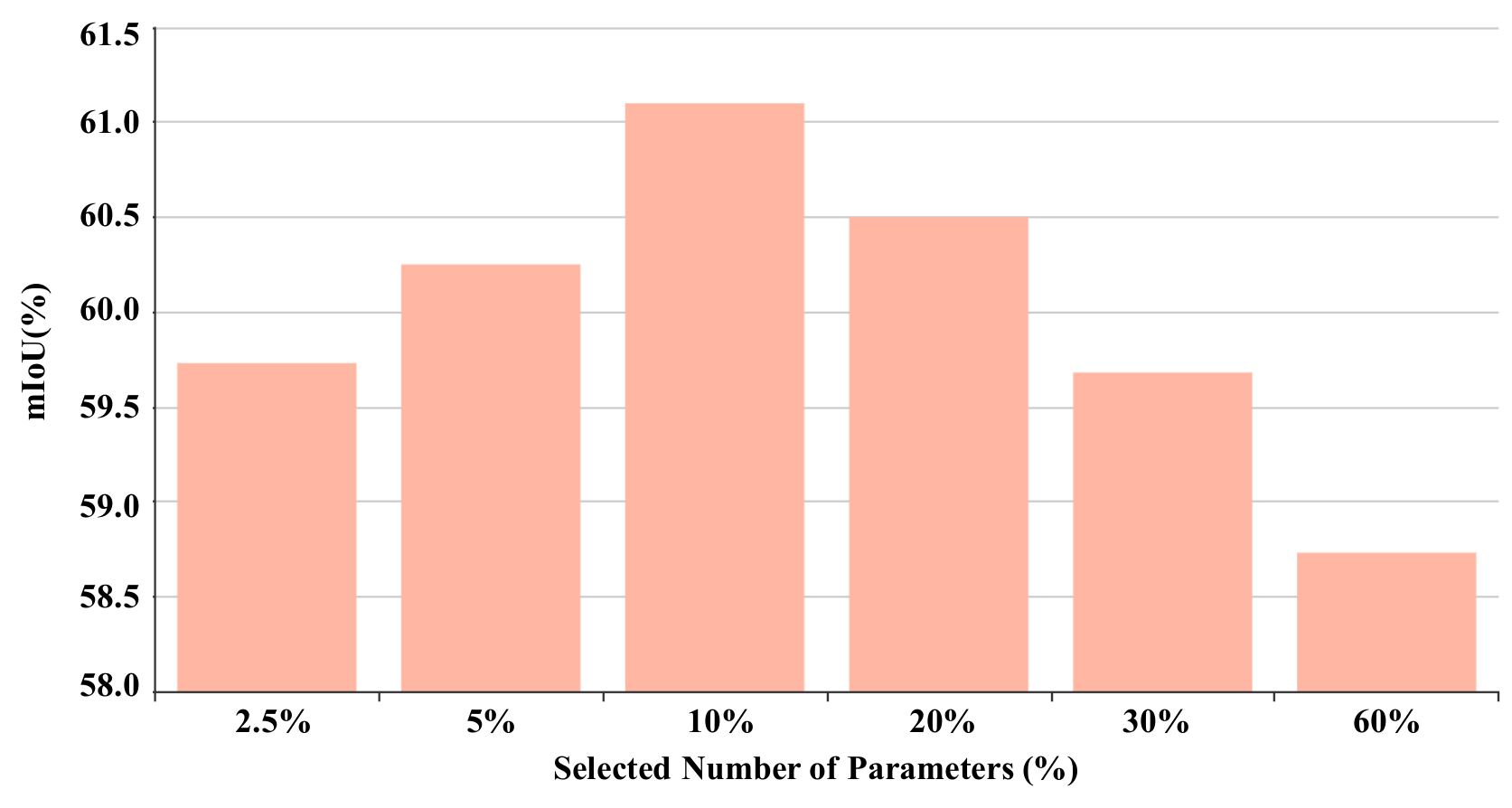}
\centering
\vspace{-0.5cm}
\caption{Different percentages of updated parameters (DSP
+ TRP) }
\label{fig:percent}
\vspace{-0.5cm}
\end{figure}

\textbf{Percentage of model updated parameters.}
We conduct an analysis of the impact of the percentage of parameters (DSP and TRP) updated by the DAT method on the CTTA scenario. The percentage of the updated parameters is controlled by the selected parameters (i.e., 0.1\%, 0.5\%, 1.0\%) for each sample. As shown in Fig.~\ref{fig:percent}, we gradually increase the percentage of parameters and record the corresponding mIoU values. We find that the mIoU initially improved with the increasing percentage of parameters, but a decline set in when the percentage surpassed 10\%.
This observation suggests that when the updated parameters are too few, they fail to capture the domain-specific knowledge effectively due to the limited number of parameters. Conversely, if there are too many updated parameters, the model undergoes excessive parameter changes, resulting in catastrophic forgetting.
Therefore, it is crucial to strike a balance on the percentage of update parameters, and we consider that DAT can achieve optimal potential in 10\%  percentage parameters.

\begin{table}[htbp]
\caption{\label{tab:Val} \textbf{Effect of select parameters' selection method.} w/o PAU means the selection of 10\% parameters from the first frame in the target sequence. w PAU denotes the utilization of our PAU method to select around 10\% parameters.}
\centering
\vspace{-0.3cm}
\setlength\tabcolsep{10pt}
\begin{adjustbox}{width=0.8\linewidth,center=\linewidth}
\begin{tabular}{c|cc|cc}
\hline
\multicolumn{1}{c|}{}   & \multicolumn{2}{c|}{w/o PAU}     & \multicolumn{2}{c}{w PAU} \\ \cline{1-5}
Method & mIoU & ACC & mIoU & ACC \\ \hline
Confidence  & 59.1 & 71.1 & 60.6 & 71.3 \\
Uncertainty  & 60.3 & 71.4 & 61.1& 72.2 \\
\hline
\end{tabular}
\end{adjustbox}
\vspace{-0.1cm}
\label{tab:select}
\end{table}

\textbf{Method of select parameters.} 
As shown in Table~\ref{tab:select}, compared to using the uncertainty method for parameter selection, select parameters based on confidence, the mIoU and ACC are significantly lower when using or not using PAU. This further demonstrates that uncertainty can effectively measure domain shift and, consequently, guide parameter selection.
Additionally, we observe that in comparison to not employing the PAU strategy, utilizing PAU can further enhance the stability of model updates, subsequently improving model performance. When using PAU and selecting parameters based on uncertainty, the model achieves mIoU and mAcc values of 61.1 and 72.2, respectively.



\section{Conclusion and Limitation}
Our Distribution-Aware Tuning (DAT) method presents an efficient and practical solution for Semantic Segmentation CTTA in real-world scenarios. DAT intelligently selects and fine-tunes two distinct sets of trainable parameters: domain-specific parameters (DSP) and task-relevant parameters (TRP). This helps mitigate issues of error accumulation and catastrophic forgetting during the continual adaptation process.
Considering the temporal nature of CTTA in autonomous driving, we propose a Parameter Accumulation Update (PAU) strategy to improve the stability of parameter selection and updates.
However, since we are addressing pixel-wise segmentation CTTA tasks, we have followed the image augmentation method outlined in \cite{Wangetal2022} to enhance accuracy. This augmentation comes with additional computational costs, which we aim to explore in future work.

\section{Acknowledgement}
Shanghang Zhang is supported by the National
Key Research and Development Project of China
(No.2022ZD0117801). This work is also supported by CCF-Baidu Open Fund Project.

\clearpage
{
\bibliographystyle{IEEEtran}
\bibliography{IEEEabrv,reference}
}

\end{document}